\documentclass[letterpaper]{article} 
\usepackage[preprint]{aaai2027} 
\usepackage[hyphens]{url} 
\usepackage{graphicx} 
\usepackage{natbib} 
\usepackage{caption} 
\usepackage{amsmath}
\usepackage{amssymb}
\usepackage{booktabs}
\usepackage{multirow}
\usepackage{makecell}
\usepackage{enumitem}
\usepackage{pifont}
\usepackage{xspace}
\usepackage{xr}
\usepackage{fontawesome}

\makeatletter
\@ifundefined{XR@@nocite}
  {\let\reedit@bibcite\bibcite
   \let\bibcite\@gobbletwo
   \externaldocument{supplementary}
   \let\bibcite\reedit@bibcite}
  {\externaldocument[][nocite]{supplementary}}
\makeatother

\makeatletter
\DeclareRobustCommand\onedot{\futurelet\@let@token\@onedot}
\def\@onedot{\ifx\@let@token.\else.\null\fi\xspace}

\makeatother

\newcommand{\projectlink}[2]{%
  \leavevmode
  \pdfstartlink attr{/Border [0 0 0]} user{%
    /Subtype /Link /A << /S /URI /URI (\pdfescapestring{#1}) >>}%
  #2\pdfendlink%
}

\title{LayerRecall: A State-Conditioned Memory Router for \\ Long-Horizon Consistency in Video Generation}

\author{
    Yixuan Ding\textsuperscript{\rm 1},
    Jiahao Kong\textsuperscript{\rm 1},
    Wei Huang\textsuperscript{\rm 2},
    Ruijie Quan\textsuperscript{\rm 1}\corresponding,
    Yi Yang\textsuperscript{\rm 1}
}
\affiliations{
    \textsuperscript{\rm 1}Zhejiang University
    \qquad
    \textsuperscript{\rm 2}The University of Hong Kong\\[0.35em]
    {\footnotesize
    \faGlobe\hspace{0.2em}
    \projectlink{https://yixuan-ding-zju.github.io/LayerRecall_Web/}{\texttt{Webpage}}
    \qquad
    \faGithub\hspace{0.2em}
    \projectlink{https://github.com/Yixuan-Ding-ZJU/LayerRecall}{\texttt{GitHub}}
    \qquad
    \raisebox{-0.25em}{\includegraphics[height=1.2em]{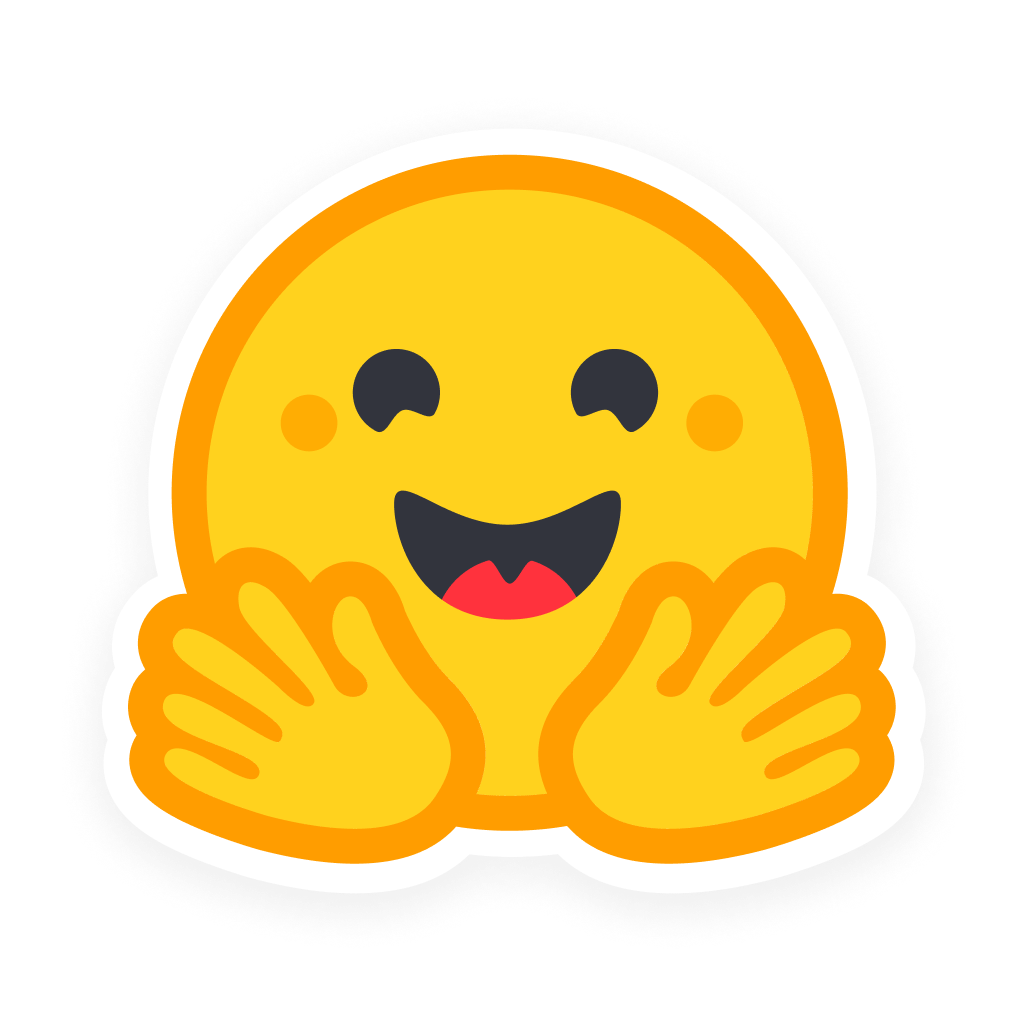}}\hspace{0.2em}
    \projectlink{https://huggingface.co/Yixuan-Ding-ZJU/LayerRecall}{\texttt{Hugging Face}}
    }
}

\newcommand{\reeditteaser}{%
  \vspace{-20pt}
  \noindent\parbox{\textwidth}{%
    \centering
    \includegraphics[width=1\textwidth,height=0.45\textheight]{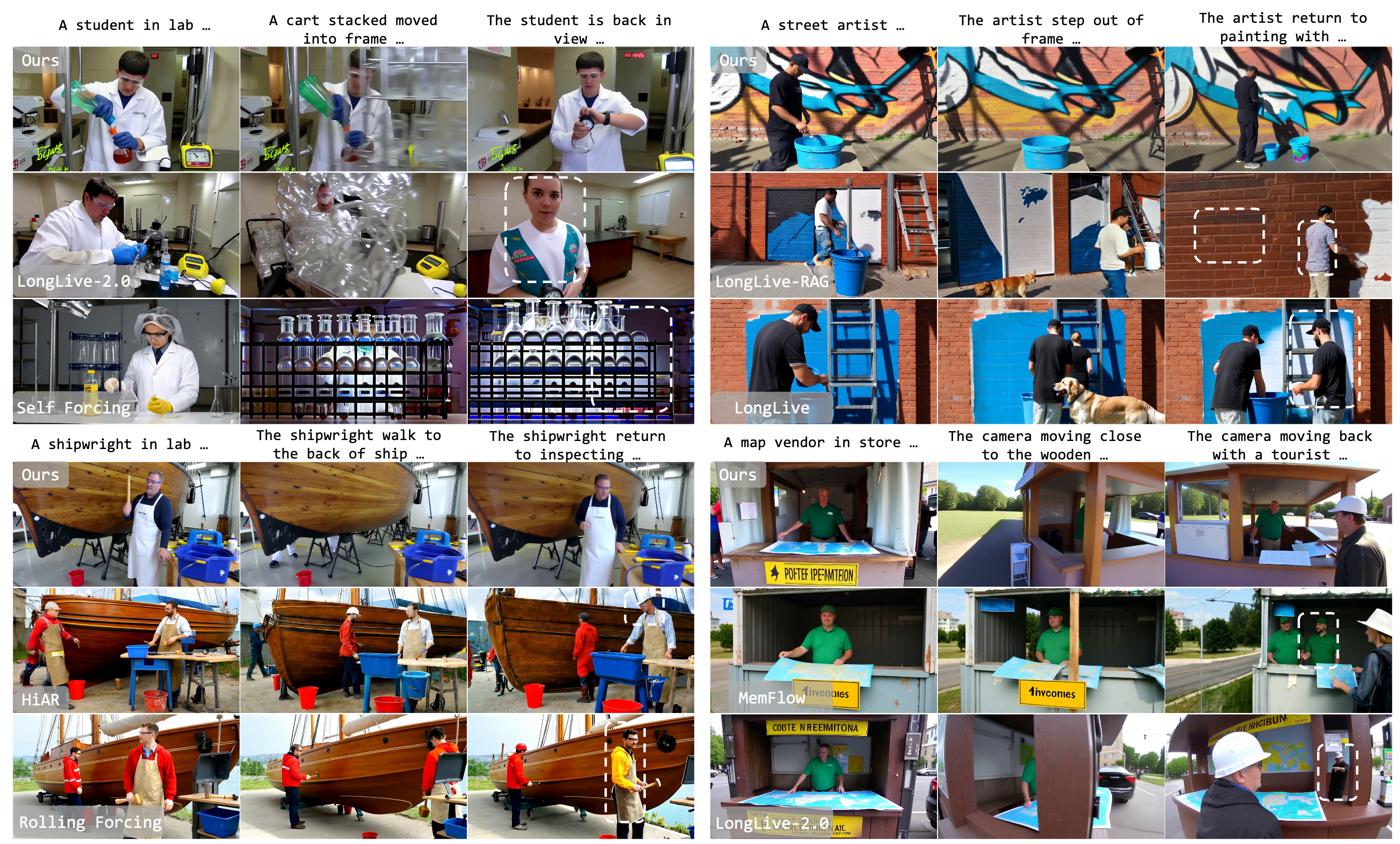}\par
    \captionsetup{skip=1pt}
  \captionof{figure}{Qualitative comparison on long-range reappearance scenarios. In each case, prompt-highlighted subjects, attributes, and objects leave the local context and later return. Compared with representative long-video baselines, LayerRecall more faithfully recalls and restores subject identity, attributes, and scene-specific objects.}
  \label{fig:teaser}
    \vspace{12pt}
  }%
}

\makeatletter
\patchcmd{\@maketitle}
  {\vskip 0.1in plus 0.5fil minus 0.05in}
  {\vskip 20pt}
  {}
  {\PackageError{main-arxiv}{Unable to adjust the first title-author spacing instance}{}}
\patchcmd{\@maketitle}
  {\vskip 0.1in plus 0.5fil minus 0.05in}
  {\vskip 20pt}
  {}
  {\PackageError{main-arxiv}{Unable to adjust the second title-author spacing instance}{}}
\patchcmd{\@maketitle}
  {{\Large{\textbf{\aaai@theauthors\ifhmode\\\fi}}}}
  {{\large\aaai@theauthors\ifhmode\\\fi}}
  {}
  {\PackageError{main-arxiv}{Unable to adjust the first author style instance}{}}
\patchcmd{\@maketitle}
  {{\Large{\textbf{\aaai@theauthors\ifhmode\\\fi}}}}
  {{\large\aaai@theauthors\ifhmode\\\fi}}
  {}
  {\PackageError{main-arxiv}{Unable to adjust the second author style instance}{}}
\patchcmd{\maketitle}
  {\twocolumn[\@maketitle]}
  {\twocolumn[\@maketitle\reeditteaser]}
  {}
  {\PackageError{reedit}{Unable to integrate the authorized teaser into the AAAI title block}{}}
\makeatother

\begin{document}

\maketitle

\begin{abstract}

Autoregressive video diffusion enables scalable long-video generation by producing chunks from a bounded recent context. While recency-based caching preserves local continuity, it evicts historical cues needed when subjects, objects, scenes, or attributes reappear. Existing memory mechanisms expose models to nonlocal history, but access alone does not ensure effective use. Our analysis reveals that video DiT layers exhibit distinct preferences for current, recent, and distant context, suggesting that long-range memory requires deciding both \emph{what to retrieve} and \emph{where to use it}. We introduce \textbf{LayerRecall, a current-conditioned, layer-selective memory router} that retrieves relevant historical K/V states and injects them only into backbone-specific memory-sensitive layers while preserving local attention elsewhere. To reduce reliance on scarce high-quality long-horizon videos and explicit memory-allocation labels, we further propose \textbf{Cross-Horizon Prediction Matching (CHPM)}, which uses a privileged long-context reference to supervise the bounded-memory router in prediction space. Across 100 multi-shot evaluation prompts, LayerRecall achieves the best overall results on MemoBench and MovieBench while matching its backbone on VBench-Long, demonstrating stronger long-range recovery without sacrificing local continuity. Qualitative analyses further reveal memory-guided self-correction, whereby initially mismatched local attributes return to their historical appearance without resetting ongoing motion or scene structure. Additional analyses show cross-backbone portability and negligible inference overhead.
\end{abstract}

\newpage
\section{Introduction}

Diffusion Transformers (DiTs)~\cite{peebles2023dit} have driven recent progress in video generation~\cite{ma2025latte,yang2025cogvideox,kong2025hunyuanvideo,wanteam2025wan}, providing a scalable backbone for high-fidelity synthesis. Most pretrained video DiTs nevertheless use bidirectional attention over a fixed clip, requiring the sequence to be processed jointly and limiting efficient extension to longer durations. Recent studies have therefore recast video diffusion as causal or autoregressive generation, producing frames or chunks from preceding context and using key-value (KV) caching to avoid recomputing the complete sequence~\cite{yin2025causvid,gao2025ca2vdm,huang2025selfforcing,yang2026longlive,liu2026rollingforcing}. These developments support streaming inference and generation beyond the duration of conventional clip-based models. As the attainable horizon grows, preserving useful information across that horizon becomes more challenging.

Longer output, however, does not by itself ensure long-term semantic consistency. Under bounded memory, autoregressive generators typically retain a recent temporal window or a fixed-size KV cache~\cite{kim2024fifodiffusion,gao2025ca2vdm,yang2026longlive,chen2026longlive2}. This recency-based policy supports local motion and appearance continuity, but eventually evicts evidence about subjects, objects, scenes, or attributes that reappear after shot changes. Existing methods augment the local pathway with fixed anchors, compressed or recurrent states, online memories, or retrieved historical frames and latents~\cite{henschel2025streamingt2v,chen2026sanavideo,yu2025videossm,hong2025slowfastvgen,yu2025contextasmemory,dou2026slotmemory,hu2026longliverag}, thereby exposing the model to nonlocal history. Yet access alone does not guarantee effective use: Figure~\ref{fig:teaser} shows that representative long-video methods still exhibit identity, attribute, count, and scene drift when content leaves the local context and later returns.

\begin{figure}[!t]
\centering
\includegraphics[width=1\columnwidth]{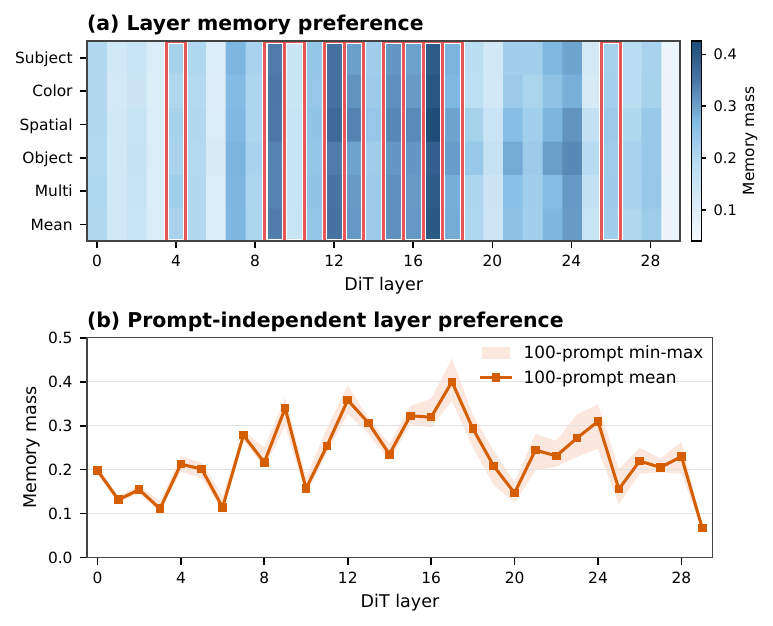}
\caption{Attention profiling on the LongLive-2.0 backbone. (a) Memory-attention mass by prompt type and DiT layer; memory denotes far-history and recent-history buckets before the current chunk. Final memory-sensitive layers are outlined. (b) Prompt-level mean and min--max memory mass across 100 prompts, showing that the layer-preference pattern remains relatively stable over the evaluated prompts.}
\label{fig:layer_preference}
\end{figure}

To understand this gap, we examine how autoregressive video DiTs distribute temporal attention across network depth. Figure~\ref{fig:layer_preference} reveals pronounced layer-wise preferences for current, recent, and distant context. This heterogeneity appears across the tested backbones, although the specific memory-sensitive layers remain backbone-dependent. Therefore, we consider that effective use of long-range memory may require not only \emph{what to retrieve}, but also \emph{where to use it}. We introduce \textbf{LayerRecall}, a current-conditioned, layer-selective memory router that jointly controls historical retrieval and injection while retaining the local pathway. Because long-horizon training videos and explicit memory-allocation labels are scarce, we further propose \textbf{Cross-Horizon Prediction Matching (CHPM)}, which turns a privileged long-context reference into prediction-space supervision for the bounded-memory router. Across 100 evaluation prompts, LayerRecall achieves SOTA results on both memory-oriented benchmarks while matching its backbone on VBench-Long, demonstrating stronger long-range recovery without sacrificing consistency and motion quality.

Our contributions are summarized as follows:
\begin{itemize}
    \item We identify layer-wise temporal preferences in autoregressive video DiTs and formulate bounded-memory use as joint allocation over historical content and network depth.
    \item We propose LayerRecall, which retrieves history from the current generation state and routes its K/V states through backbone-specific memory-sensitive layers while preserving local attention elsewhere.
    \item We propose CHPM, which uses long-context predictions to supervise sparse memory routing without explicit memory-allocation labels or backbone finetuning.
    \item We evaluate LayerRecall across multi-shot consistency, temporal stability, cross-backbone portability, and inference efficiency.
\end{itemize}



\begin{figure*}[!t]
\centering
\includegraphics[width=1\textwidth]{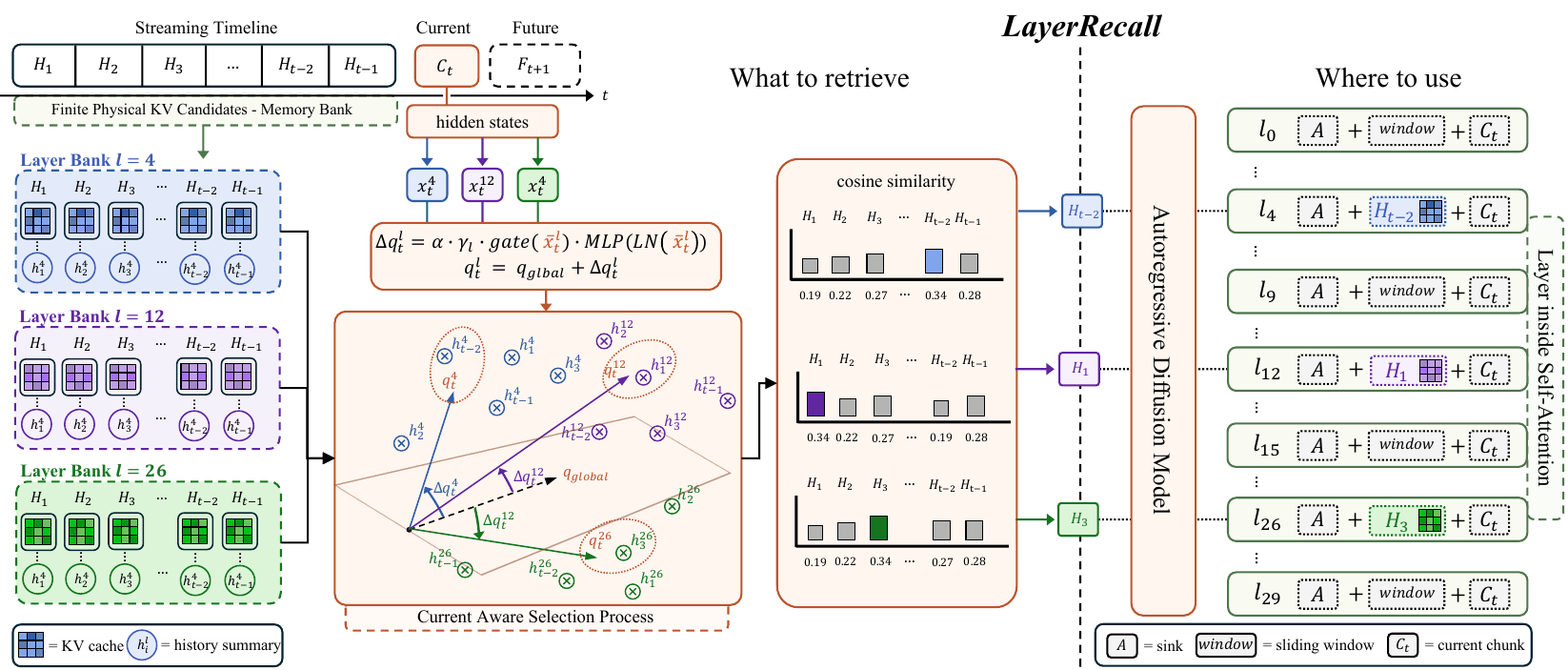}
\caption{Overview of LayerRecall. Given the current chunk, the router retrieves finite historical K/V candidates from layer-indexed memory banks and scores which memories are relevant to the next generation step. Retrieved states are injected only into profiled memory-sensitive layers, while the remaining layers keep the original sink, sliding-window, and current-chunk attention context. K/V denotes key-value states.}
\label{fig:workflow}
\end{figure*}

\section{Related Works}

\paragraph{Autoregressive Video Generation.}
Diffusion Transformers~\cite{peebles2023dit} have become a common backbone for high-quality video generation, but conventional bidirectional formulations typically process a fixed clip as a whole. Recent work has recast video diffusion as causal or autoregressive generation, producing frames or chunks sequentially and reusing cached context to support streaming inference and longer rollouts~\cite{yin2025causvid,gao2025ca2vdm,huang2025selfforcing,yang2026longlive,liu2026rollingforcing}. Follow-up studies further address robustness to self-generated histories, long-horizon inference, and causal training~\cite{cui2026selfforcingpp,zhu2026causalforcing}. Together, these methods establish an efficient foundation for autoregressive long-video generation, while primarily addressing causal conversion, rollout robustness, and generation efficiency rather than selective access to distant historical evidence.

\paragraph{Long-Horizon Video Generation.}
One line of work extends pretrained short-video models through noise rescheduling, overlapping windows, queue-based generation, or global-local information exchange~\cite{qiu2024freenoise,kim2024fifodiffusion,lu2024freelong,li2025longdiff}. Streaming systems similarly rely on bounded local context or fixed anchors to preserve continuity across successive segments~\cite{henschel2025streamingt2v,yang2026longlive,liu2026rollingforcing}. A complementary direction enlarges available temporal context or trains autoregressive models to manage longer histories~\cite{guo2025longcontexttuning,chen2026contextforcing}. These approaches substantially extend generation horizons, but local propagation and fixed anchors may not preserve sparse, content-dependent evidence after it leaves the active window, whereas dense long-context processing increases computation and memory costs.

\paragraph{Memory-Augmented Video Generation.}
Recent methods preserve information beyond the local window through compressed global states~\cite{chen2026sanavideo,yu2025videossm}, online parameter memories or memory experts~\cite{hong2025slowfastvgen,stapf2026memoryexperts}, and selective retention or retrieval of historical frames, latents, or key-value states~\cite{yu2025contextasmemory,yi2026deepforcing,dou2026slotmemory,hu2026longliverag}. These directions demonstrate that nonlocal history can be represented and queried without exposing the complete sequence at every generation step. They also motivate a complementary question: beyond deciding what history to retrieve, where should that history enter the network without disrupting local temporal modeling? LayerRecall addresses these two dimensions jointly by retrieving history from the current generation state and routing it only to memory-sensitive layers identified through layer-wise empirical analysis, while preserving local attention elsewhere.

\section{Method}

 We formulate effective historical access as two decisions: \emph{what to retrieve}, which assigns the available memory slots to content-relevant chunks, and \emph{where to use it}, which determines the network layers that receive the retrieved K/V states. \textbf{LayerRecall} implements these decisions through current-conditioned retrieval and layer-selective routing. We train it without explicit memory-selection labels using \textbf{Cross-Horizon Prediction Matching (CHPM)}, which transfers the predictive behavior of a privileged long-context reference to a bounded-memory student.

\begin{figure*}[!t]
\centering
\includegraphics[width=1\textwidth]{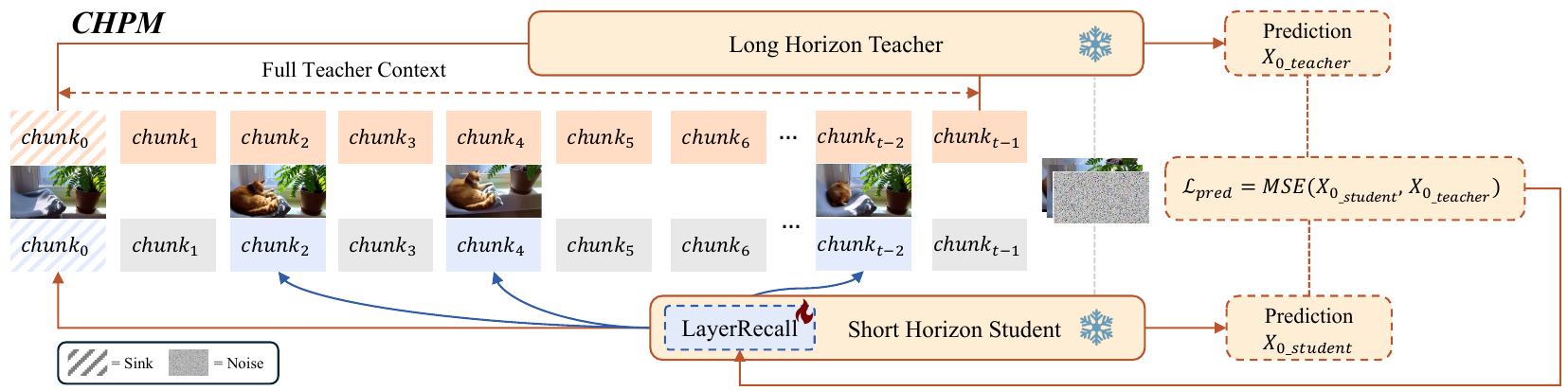}
\caption{Training framework of LayerRecall. A frozen long-horizon teacher predicts from an expanded historical context, while a frozen short-horizon student predicts with bounded local context augmented by LayerRecall memory routing. The routing module is optimized by matching student and teacher predictions; snowflakes indicate frozen backbones.}
\label{fig:training_pipeline}
\end{figure*}

\subsection{LayerRecall}

Figure~\ref{fig:workflow} gives an overview. LayerRecall preserves the local pathway of a pretrained chunk-autoregressive DiT while adding sparse access to historical K/V states.

\paragraph{What to Retrieve: Current-Conditioned Historical Retrieval.}

During each context update, every layer pools the normalized pre-RoPE keys of a historical chunk into a detached summary $m_i^l$ and stores it with chunk metadata and a pointer to the corresponding physical-cache slice. The summary is used only for scoring; the full cached RoPE-applied keys and corresponding values are the payload that enters attention after selection. The physical K/V cache follows a sink-protected sliding update: the sink is retained, while the oldest non-sink states are removed as new chunks are appended.

For the current chunk, LayerRecall mean-pools the detached pre-attention hidden state at layer $l$ into $c_t^l$ and sets $z_t^l=\mathrm{LN}(c_t^l)$. It combines a shared learnable query with a current-conditioned residual:
\begin{equation}
q_t^l
=
q_{\mathrm{global}}
+
\alpha\,\gamma_l\,
\sigma\!\left(W_g z_t^l+b_g\right)
\mathrm{MLP}(z_t^l).
\label{eq:current_query}
\end{equation}
The normalization, MLP, gate, global query, and scalar $\alpha$ are shared across layers, while $\gamma_l$ is layer-specific. Each layer independently ranks physically resident historical records by the cosine similarity between $q_t^l$ and $m_i^l$. During training, every retrieval slot uses the highest-scoring chunk's full K/V in the forward pass and a temperature-controlled soft mixture as a straight-through gradient surrogate; later slots exclude chunks already selected. Thus, the attention computation remains hard and budgeted, while the selector receives differentiable learning signals.

\paragraph{Where to Use: Layer-Selective Memory Injection.}

Figure~\ref{fig:layer_preference} shows that temporal attention preferences vary substantially across layers. We identify a fixed, backbone-specific set $\mathcal L_{\mathrm{mem}}$ through temporal profiling, routing-usage analysis, and controlled policy comparisons; the full protocol is provided in the supplementary material. Layers in $\mathcal L_{\mathrm{mem}}$ attend to the sink, hard-selected historical K/V slots, and the current chunk. All remaining layers retain the backbone's original sink-plus-sliding-window context and do not inject retrieved history. Consequently, \emph{what to retrieve} is current-dependent, whereas \emph{where to use it} is a fixed backbone-specific policy.

\subsection{CHPM: Cross-Horizon Prediction Matching}

CHPM trains LayerRecall to approximate long-context predictive behavior under a bounded memory budget, without requiring high-quality long-horizon target videos or explicit memory-allocation labels. As illustrated in Figure~\ref{fig:training_pipeline}, teacher and student share the same pretrained causal video backbone, whose parameters remain frozen; only the LayerRecall parameters $\phi$ are optimized. Their distinction is the available history: the teacher directly attends to an expanded context, whereas the student predicts from its bounded local context augmented by LayerRecall.

\paragraph{Cross-Horizon Reference and Prediction Matching.}

At each supervised anchor $a\in\mathcal A$, teacher and student receive the same current noisy latent, prompt condition, and diffusion timestep, while maintaining independent rollout histories and K/V caches. The teacher follows its own no-grad long-context trajectory and provides a detached denoised-latent prediction $\widehat{x}_{0,a}^{T}$. The student follows its bounded-memory trajectory and produces $\widehat{x}_{0,a}^{S}$. The teacher therefore serves as a privileged-context behavioral reference rather than a different or intrinsically stronger generative model.

We define the prediction-matching term over supervised anchors as
\begin{equation}
\mathcal L_{\mathrm{pred}}
=
\frac{1}{|\mathcal A|}
\sum_{a\in\mathcal A}
\left\|
\widehat{x}_{0,a}^{S}
-
\mathrm{sg}\!\left(\widehat{x}_{0,a}^{T}\right)
\right\|_2^2
.
\label{eq:chpm_pred}
\end{equation}
The complete training objective is
$\mathcal L_{\mathrm{CHPM}}=\lambda_{\mathrm{pred}}\mathcal L_{\mathrm{pred}}+
\lambda_{\mathrm{reg}}\mathcal L_{\mathrm{reg}}$, where
$\mathcal L_{\mathrm{reg}}$ averages the mean-squared magnitude of each
trainable LayerRecall parameter tensor. This regularization is applied only to
the router; the frozen generative backbone is excluded. Full loss settings are
provided in the supplementary material.
CHPM does not observe which historical chunk the teacher uses and does not match its attention distribution. Instead, with both backbones frozen, reducing the prediction gap requires LayerRecall to change how historical K/V states are retrieved and injected. The prediction discrepancy thus converts the teacher's long-context behavior into task-relevant supervision for bounded-memory routing.

\paragraph{Detached Student Rollout.}

LayerRecall must be trained on the context distribution encountered during autoregressive inference. Supplying teacher-generated prefixes to the student would hide errors accumulated by the bounded-memory model and introduce a train--test mismatch. We therefore let the student maintain its own trajectory. Non-anchor chunks are denoised without gradients and written back as detached future context; supervised anchors compute the CHPM loss and are likewise detached before subsequent context updates. The teacher independently advances its own detached long-context trajectory and provides targets only at supervised anchors. This design exposes LayerRecall to the student's self-generated history while preventing gradients from spanning the entire video, keeping supervision localized to the current anchor and memory use bounded across chunks.

\begin{figure*}[!t]
\centering
    \includegraphics[width=1\textwidth]{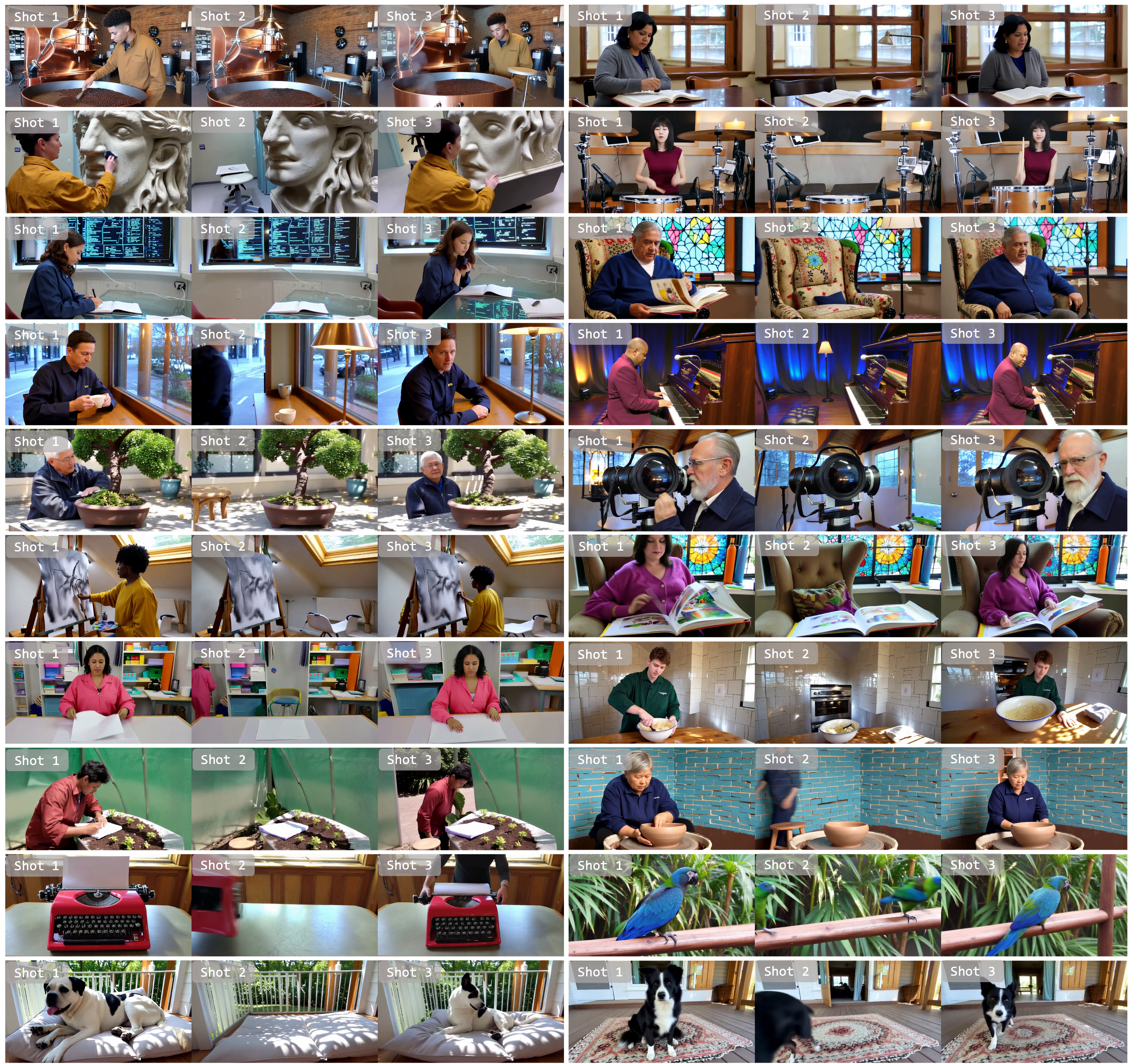}
\caption{LayerRecall showcases across diverse three-shot reappearance
scenarios. Each triplet samples one frame from Shots 1--3: a subject, object,
or visual attribute is established in Shot 1, leaves the visible scene in Shot
2, and reappears in Shot 3. Across people, animals, apparel, instruments, and
scene-specific objects, LayerRecall preserves the corresponding identity and
fine-grained appearance over the intervening temporal gap, demonstrating
stable long-term recall.}
\label{fig:quality_comparison}
\end{figure*}

\begin{table*}[!t]
\centering
\small
\setlength{\tabcolsep}{2.4pt}
\renewcommand{\arraystretch}{1.2}
\begin{tabular*}{\textwidth}{@{\extracolsep{\fill}}lcccccccccc@{}}
\toprule
\multicolumn{3}{c}{} & \multicolumn{4}{c}{MemoBench $\uparrow$} & \multicolumn{4}{c}{MovieBench $\uparrow$} \\
\cmidrule(lr){4-7}\cmidrule(l){8-11}
Method & Venue & \makecell{VBench-Long\\Avg.} & Overall & \makecell{Obj.\\Reapp.} & \makecell{Layout\\Rec.} & \makecell{State\\Upd.} & Overall & \makecell{Shot\\Coh.} & \makecell{Scene\\Layout} & Narrative \\
\midrule
LongLive & ICLR26 & 0.977 & 0.499 & 0.516 & 0.534 & 0.464 & 0.544 & 0.491 & 0.624 & 0.509 \\
LongLive-2.0 & arXiv26 & 0.978 & 0.513 & \underline{0.519} & 0.524 & \underline{0.499} & 0.546 & \underline{0.531} & 0.596 & \underline{0.517} \\
LongLive-RAG & arXiv26 & 0.958 & 0.373 & 0.358 & 0.360 & 0.297 & 0.394 & 0.352 & 0.436 & 0.355 \\
CausVid & CVPR25 & 0.975 & 0.426 & 0.418 & 0.414 & 0.400 & 0.483 & 0.489 & 0.495 & 0.450 \\
Deep Forcing & ICML26 & 0.964 & 0.408 & 0.371 & 0.441 & 0.405 & 0.511 & 0.473 & 0.583 & 0.427 \\
Dummy Forcing & arXiv26 & 0.953 & 0.417 & 0.427 & 0.410 & 0.407 & 0.494 & 0.480 & 0.486 & 0.502 \\
Self-Forcing & NeurIPS25 & 0.976 & 0.327 & 0.316 & 0.346 & 0.293 & 0.363 & 0.292 & 0.371 & 0.344 \\
MemFlow & arXiv25 & \underline{0.981} & \underline{0.531} & 0.516 & \textbf{0.573} & 0.492 & 0.542 & 0.480 & 0.630 & 0.504 \\
Rolling Forcing & ICLR26 & 0.972 & 0.482 & 0.498 & 0.483 & 0.481 & \underline{0.548} & 0.505 & 0.608 & 0.510 \\
Context Forcing & ICML26 & 0.971 & 0.417 & 0.421 & 0.455 & 0.401 & 0.463 & 0.421 & 0.497 & 0.440 \\
HiAR & arXiv26 & 0.969 & 0.424 & 0.347 & 0.454 & 0.429 & 0.495 & 0.364 & \underline{0.646} & 0.385 \\
Infinity-RoPE & CVPR26 & 0.978 & 0.455 & 0.408 & 0.536 & 0.477 & 0.513 & 0.446 & 0.589 & 0.445 \\
SkyReels-V2 & arXiv25 & \textbf{0.992} & 0.466 & 0.404 & 0.489 & 0.463 & 0.508 & 0.400 & \textbf{0.665} & 0.380 \\
\midrule
\textbf{LayerRecall} & - & 0.978 & \textbf{0.548} & \textbf{0.571} & \underline{0.572} & \textbf{0.525} & \textbf{0.578} & \textbf{0.579} & 0.619 & \textbf{0.561} \\
\bottomrule
\end{tabular*}

\caption{Main results on the videos generated from 100 evaluation prompts. VBench-Long Avg. is the mean of VBench-Long subject consistency, background consistency, and motion smoothness. MemoBench and MovieBench provide task-aligned memory diagnostics. Bold and underline denote the best and second-best scores.}
\label{tab:main_quality_quantity}
\end{table*}

\begin{figure}[!t]
\centering
\includegraphics[width=1\columnwidth]{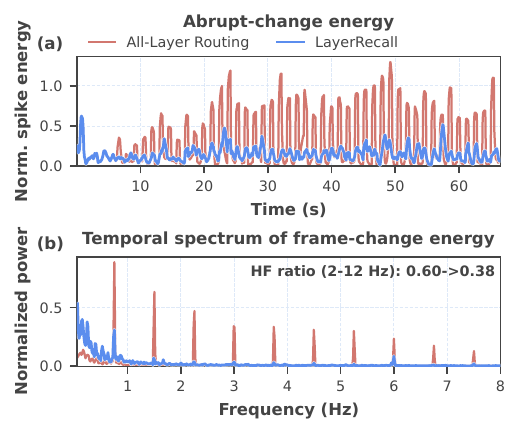}
{\captionsetup{type=figure}
\caption{Temporal stability analysis on the 100 evaluation prompts. (a) Abrupt-change energy over time, with fewer frame-change spikes under LayerRecall. (b) Temporal spectrum of frame-change energy, with the HF ratio reduced from 0.60 to 0.38.}
\label{fig:flicker_spectrum}}
\end{figure}

\section{Experiments}
\subsection{Experimental Setups}

\paragraph{Implementation details.}
We build LayerRecall upon LongLive-2.0~\cite{chen2026longlive2}, whose backbone follows the Wan2.2-TI2V-5B architecture~\cite{wanai2025wan22ti2v5b}. Each training sequence contains 384 latent frames, partitioned into 48 chunks of 8 frames, and CHPM uses one denoising step during training. The teacher attention horizon is full 384 frames context (48 chunks), whereas the student operates under a 32-frame attention-visible budget and maintains an 80-frame physical K/V cache (10 chunks). At each memory-sensitive layer, this budget comprises an 8-frame sink, up to two hard-selected historical chunks (16 latent frames), and the current 8-frame chunk; the memory-sensitive layers have 0-based indices $[4,9,10,12,13,15,16,17,18,26]$, while the remaining 20 layers preserve the original local sliding-window attention. We train on 16 NVIDIA H100 GPUs using sequence parallelism of 2 and data parallelism of 8 for one epoch.

\begin{table}[t]
\centering
\setlength{\tabcolsep}{3pt}
\small

\begin{tabular}{p{0.4\columnwidth}ccc}
\toprule
Metric & \shortstack{All-Layer\\Routing} & LayerRecall & Diff. \\
\midrule
VBench Motion Smooth. $\uparrow$ & 0.9910 & \textbf{0.9916} & +0.0006 \\
Helios Smooth. $\uparrow$ & 0.9917 & \textbf{0.9923} & +0.0006 \\
CLIP Adjacent $\uparrow$ & 0.9397 & \textbf{0.9574} & +0.0177 \\
DINO Adjacent $\uparrow$ & 0.8795 & \textbf{0.9212} & +0.0417 \\
\bottomrule
\end{tabular}

\caption{Temporal consistency ablation between LayerRecall and all-layer memory routing.}
\label{tab:layer_vs_all}
\end{table}

\paragraph{Data Construction.}
We train LayerRecall on 1,600 multi-shot prompts, each containing 48 blocks aligned with the 384-frame training rollout. We separately use 100 prompts to generate videos for evaluation. Prompt construction, filtering, and single-prompt adaptation for methods without a native multi-shot interface are described in the supplementary material.

\paragraph{Baselines.}
We compare with autoregressive long-video generators, including CausVid, Deep Forcing, Dummy Forcing, LongLive, LongLive-2.0, Self-Forcing, Rolling Forcing, and SkyReels-V2~\cite{yin2025causvid,yi2026deepforcing,guo2026dummyforcing,yang2026longlive,chen2026longlive2,huang2025selfforcing,liu2026rollingforcing,chen2025skyreelsv2}. We also include memory- or context-augmented methods: LongLive-RAG, MemFlow, Context Forcing, HiAR, and Infinity-RoPE~\cite{hu2026longliverag,ji2025memflow,chen2026contextforcing,zou2026hiar,yesiltepe2026infinityrope} for more comprehensive comparisons.

\paragraph{Evaluation metrics.}
We assess long-range consistency with subject consistency, background consistency, and motion smoothness of VBench-Long, together with the memory-oriented MemoBench and MovieBench scores~\cite{huang2024vbench,huang2026vbenchpp,chen2026memobench,wu2025moviebench}.

\subsection{Qualitative Analysis}

Figure~\ref{fig:quality_comparison} presents LayerRecall generations across diverse three-shot sequences. In each case, Shot 1 establishes a subject, object, or visual attribute, Shot 2 removes it from the visible scene, and Shot 3 tests its recovery after the relevant evidence has fallen outside the local context. LayerRecall consistently restores identities and fine-grained appearance across people, animals, apparel, instruments, and scene-specific objects, while allowing the intervening shot to evolve independently. These cases provide direct visual evidence of stable long-term recall over substantial temporal gaps.

Beyond recovering an earlier appearance when it re-enters the scene, we repeatedly observe a \emph{memory-guided self-correction} phenomenon: an initially mismatched local attribute can return to its historical state later within the same shot, without resetting the ongoing action or scene structure. Figure~\ref{fig:sup_self_correction} and the supplementary analysis provide a representative example and discuss its connection to LayerRecall's state-conditioned retrieval and layer-selective memory injection.

\subsection{Quantitative Results}

Table~\ref{tab:main_quality_quantity} shows a consistent advantage on memory-oriented evaluation. LayerRecall ranks first on both MemoBench and MovieBench overall and remains among the top methods across most diagnostic subdimensions. It simultaneously matches LongLive-2.0 on VBench-Long, indicating that these memory gains do not compromise the backbone's consistency and motion quality. This balance is consistent with selective retrieval and layer-wise injection strengthening distant-memory use while preserving the local pathway.

\begin{figure}[!t]
\centering
\includegraphics[width=1\columnwidth]{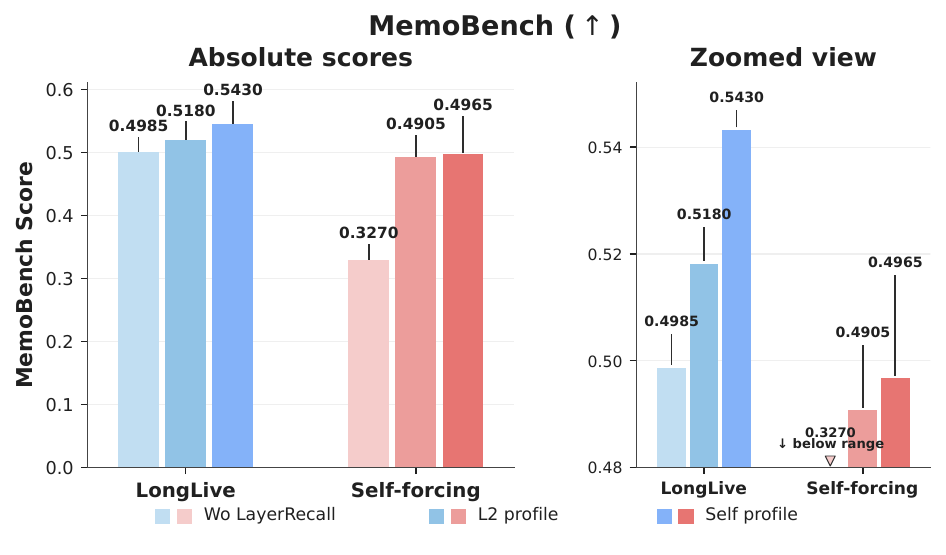}
{\captionsetup{type=figure}
\caption{MemoBench evaluation of zero-training cross-backbone portability on 100 prompts. “L2 profile” applies LayerRecall with the LongLive-2.0 layer policy, while “Self profile” uses a backbone-specific layer policy obtained by profiling each target backbone.}
\label{fig:cross_backbone_adapt}}
\end{figure}

\begin{table}[t]
\centering
\setlength{\tabcolsep}{3.2pt}

\begin{tabular}{llcc}
\toprule
Backbone & Variant & 
\shortstack{End-to-end\\time (s/video) $\downarrow$} & 
FPS $\uparrow$ \\
\midrule

\multirow{2}{*}{LongLive-2.0} 
& w/o LayerRecall & 305.9 & 5.22 \\

\cmidrule(l){2-4}

& w/ LayerRecall & 309.4 & 5.16 \\

\bottomrule
\end{tabular}

\caption{Inference efficiency under matched H100 settings. FPS is computed from decoded frames divided by full end-to-end generation time.}
\label{tab:speed_ablation}

\end{table}

\begin{figure}[!t]
\centering
\includegraphics[width=1\columnwidth]{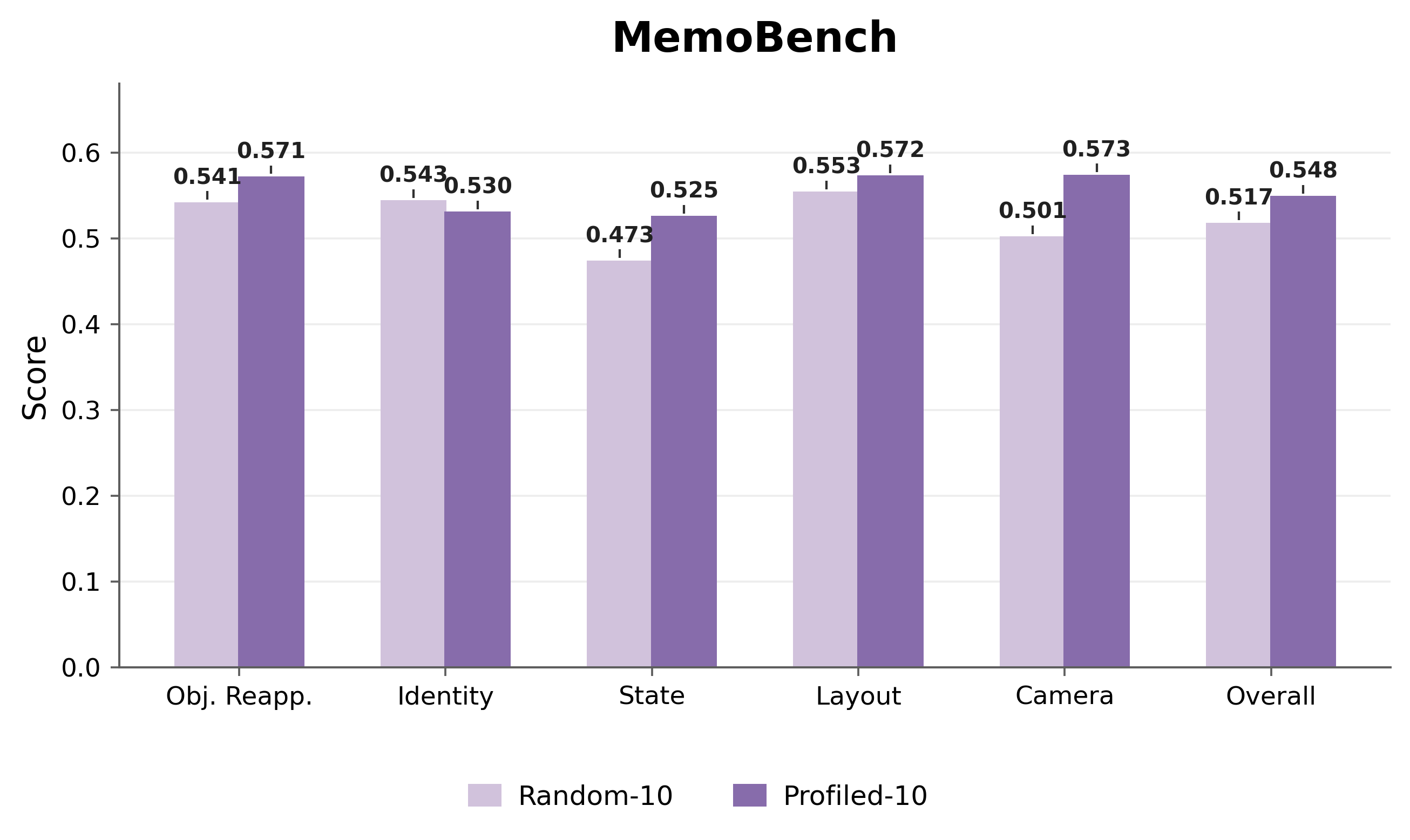}
{\captionsetup{type=figure}
\caption{MemoBench comparison of random and profile-guided historical-memory
insertion in the dedicated matched ablation over 100 evaluation prompts. Under
matched retrieval settings,
LayerRecall (Profiled-10) improves the overall score and four of five reported dimensions,
supporting the importance of selecting where retrieved memory enters the DiT.}
\label{fig:supp_layer_policy_ablation}}
\end{figure}

\begin{table}[t]
\centering
\small
\setlength{\tabcolsep}{3.2pt}

\begin{tabular}{llcccc}
\toprule
Method & Variant & \shortstack{Memo\\Overall $\uparrow$} & 
\shortstack{Memo\\ID $\uparrow$} & 
\shortstack{Memo\\State $\uparrow$} &
\shortstack{Memo\\Camera $\uparrow$}  \\
\midrule

\multirow{2}{*}{LayerRecall} 
& Random init 
& 0.519 & 0.484 & 0.487 & 0.508 \\

\cmidrule(l){2-6}

& Trained 
& \textbf{0.548} 
& \textbf{0.530} 
& \textbf{0.525} 
& \textbf{0.573} \\

\bottomrule
\end{tabular}

\caption{Ablation of CHPM validity on the MemoBench. Both variants use the same ten memory-sensitive layers.}
\label{tab:chpm_training_ablation}

\end{table}

\subsection{Ablation Study}

\subsubsection{Analysis of Efficiency}
Table~\ref{tab:speed_ablation} compares LayerRecall with the LongLive-2.0 backbone under matched H100 settings. LayerRecall changes end-to-end generation time from $305.9$ to $309.4$ seconds per video, showing that its selective memory routing introduces negligible inference overhead.









\subsubsection{Effect of Layer-Selective Memory Routing}

This ablation isolates where retrieved historical K/V states enter the generator. We compare all-layer routing with LayerRecall while keeping the checkpoint, retrieved memory, memory budget, and evaluation protocol fixed. 

Figure~\ref{fig:flicker_spectrum} and Table~\ref{tab:layer_vs_all} show that LayerRecall reduces the high-frequency frame-change power from $0.60$ to $0.38$ and improves motion smoothness and adjacent-frame CLIP/DINO consistency. These results support profile-guided routing as a means of limiting the temporal disturbance introduced by broad all-layer memory access.

\subsubsection{Backbone Portability of LayerRecall}

We test zero-training portability on LongLive~\cite{yang2026longlive} and Self-Forcing~\cite{huang2025selfforcing}, which use the 1.3B Wan2.1 backbone family~\cite{wanteam2025wan} rather than 5B Wan2.2 of LongLive-2.0~\cite{chen2026longlive2,wanai2025wan22ti2v5b}. We directly reuse all trained LayerRecall parameters without further optimization. We compare the frozen backbone alone, direct reuse with the original LongLive-2.0 layer profile, and the same reused LayerRecall with only the enabled-layer policy replaced by a 1.3B profile obtained from each model corresponding.

Figure~\ref{fig:cross_backbone_adapt} highlights the resulting MemoBench comparison, while complete metrics are provided in the supplementary material. Direct reuse improves MemoBench Overall on both backbones and most reported dimensions, supporting the portability of the state-conditioned \emph{what-to-retrieve} policy. Replacing only the layer profile yields the best Overall score on both models, although individual dimensions exhibit trade-offs, further indicating the importance of adapting \emph{where-to-use} to the target backbone.
We interpret these results as cross-backbone portability over the tested models rather than universal zero-shot applicability.

\subsubsection{Effect of CHPM}
Table~\ref{tab:chpm_training_ablation} compares LayerRecall trained with CHPM against the same architecture with randomly initialized routing parameters; both variants use the same ten memory-sensitive layers. CHPM raises MemoBench overall from $0.519$ to $0.548$ and improves every reported subdimension, supporting prediction matching as effective supervision beyond the fixed layer policy.

\subsubsection{Effect of Profile-Guided Where to Use}

We isolate the \emph{where-to-use} decision by applying the same trained
LayerRecall checkpoint to either the profiled LongLive-2.0 policy
$[4,9,10,12,13,15,16,17,18,26]$ or ten randomly sampled DiT layers
$[0,3,4,7,8,17,20,23,24,28]$. All other settings, including the retrieval
mechanism, memory budget, and the 100 evaluation prompts, remain fixed. Figure~\ref{fig:supp_layer_policy_ablation} show that the profiled policy raises the
MemoBench overall score from $0.538$ to $0.570$ and improves object
reappearance, state, layout, and camera consistency. Identity is the only
reported dimension on which Random-10 is higher ($0.507$ versus $0.495$). The overall and four-of-five dimensional
gains show that, under a fixed checkpoint and retrieval procedure, the layer
allowlist materially affects how well routed history is used. This supports
profile-guided \emph{where-to-use} routing over an equally sized random policy,
without assuming that different layers select identical historical chunks.

\section{Conclusion}
\label{sec:conclusion}
We presented LayerRecall for improving long-range consistency in autoregressive video diffusion under a bounded physical K/V cache. Motivated by layer-wise differences in temporal attention, LayerRecall revisits memory use from a layer-aware perspective and separates memory management into two coupled decisions: a current-conditioned query determines \emph{what to retrieve}, while a backbone-specific policy determines \emph{where to use} the retrieved K/V states. Retaining the original local pathway elsewhere enables distant recall without discarding local temporal evolution.
CHPM learns this routing strategy from a privileged long-context reference without explicit memory-allocation labels or backbone fine-tuning. Prediction-space matching converts long-context behavior into routing supervision, while detached self-rollout exposes the student to its own accumulated context without propagating gradients across the full sequence.

Across multi-shot evaluations, LayerRecall achieves the best overall scores on MemoBench and MovieBench while matching LongLive-2.0 on VBench-Long. Ablations validate layer-selective injection, CHPM, and cross-backbone transfer with negligible overhead. Together, these results highlight the importance of allocating limited history across both content and network depth. Future work may explore adaptive layer activation and compressed memory.

{\small
\bibliography{main}
}
\def\LayerRecallAppendixTitleRendered{}
\twocolumn[
  \begin{center}
    {\LARGE\bfseries LayerRecall: A State-Conditioned Memory Router for \\ Long-Horizon Consistency in Video Generation \\[0.5em] Appendix}
  \end{center}
  \vspace{0.5em}
]

\appendix
\ifdefined\LayerRecallAppendixTitleRendered
\else
\newpage
\begin{center}
{\LARGE\bfseries Appendix}
\end{center}
\fi
\section{Additional Experiments}
\label{sec:supp_additional_experiments}

\subsection{Layer Preference Beyond LongLive-2.0}
\label{sec:supp_cross_backbone_preference}

\begin{figure*}[t]
\centering
\begin{minipage}[t]{0.485\textwidth}
    \centering
    \includegraphics[width=\linewidth]{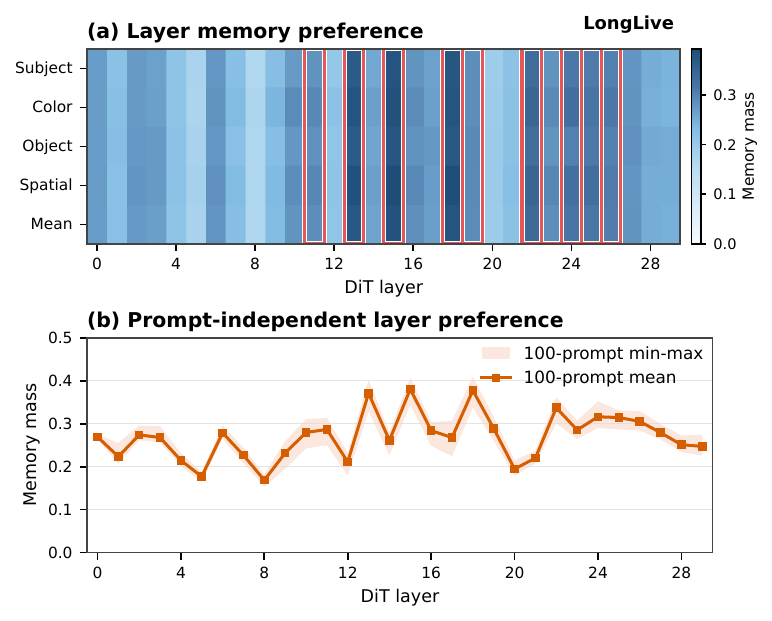}
\end{minipage}
\hfill
\begin{minipage}[t]{0.485\textwidth}
    \centering
    \includegraphics[width=\linewidth]{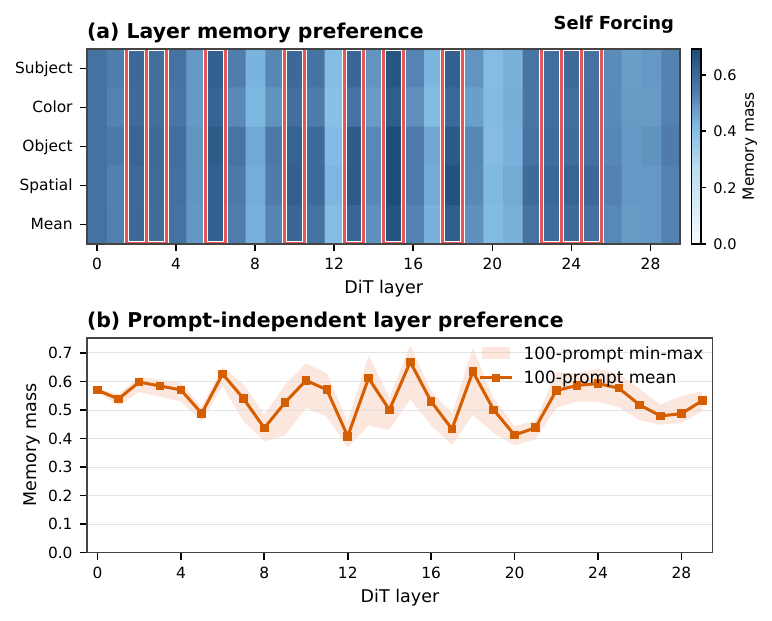}
\end{minipage}
\caption{Layer-preference profiling on LongLive (left) and Self-Forcing
(right), two backbones distinct from the LongLive-2.0 model used in the main
experiments. For each backbone, the upper panel reports historical attention
mass across DiT layers and prompt categories, while the lower panel shows the
mean and min--max range over 100 held-out prompts. Both models exhibit
heterogeneous and relatively prompt-stable layer-wise preferences, although their specific
memory-sensitive layers differ. Red outlines mark the profiled insertion
policies used in the cross-backbone experiments.}
\label{fig:supp_cross_backbone_profile25}
\end{figure*}

Figure~\ref{fig:supp_cross_backbone_profile25} extends the layer-preference
analysis in Main.Fig.2 beyond the LongLive-2.0 backbone.
We profile LongLive~\cite{yang2026longlive} and
Self-Forcing~\cite{huang2025selfforcing} on the same 100 held-out prompts and
aggregate historical attention mass with the protocol used in the main paper.
Both backbones exhibit pronounced variation across network depth: layers that
allocate more attention to historical context form heterogeneous,
model-specific patterns rather than a uniform response throughout the DiT. The relative
layer-wise trends are also stable across most evaluated prompts, as reflected
by the narrow prompt-level ranges at many layers. These observations support
the presence of layer preference beyond LongLive-2.0, while the different peak
locations indicate that the exact memory-sensitive layer set remains
backbone-specific. We regard this as evidence over the tested models and prompt
set, rather than a universal property of all video DiTs.

\subsection{Memory-Guided Self-Correction}
\label{sec:supp_self_correction}

\begin{figure*}[!t]
\centering
    \includegraphics[width=1\textwidth]{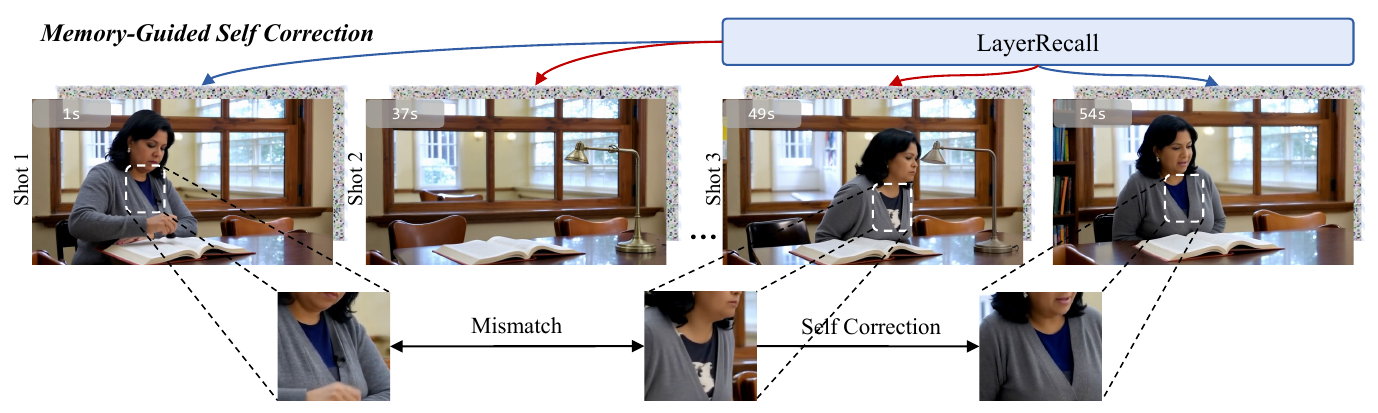}
\caption{Memory-guided self-correction in a three shot sequence. The subject
wears a solid blue inner garment in Shot 1, leaves the scene in Shot 2, and
initially reappears with a mismatched pattern in Shot 3. LayerRecall later
recalls the earlier blue garment while preserving the subject's identity,
gray cardigan, ongoing action, and scene structure, illustrating localized
attribute recovery without a global visual reset.}
\label{fig:sup_self_correction}
\end{figure*}

In addition to recovering historical attributes when a subject reappears, we
observe cases in which LayerRecall corrects an initial local mismatch later
within the same shot. Figure~\ref{fig:sup_self_correction} shows one example:
the returning subject first appears with an incorrect patterned inner garment,
then recovers the solid blue garment established before she left the scene.
The correction remains localized to the mismatched attribute, while her
identity, gray cardigan, action, and surrounding scene continue without a
global visual reset.

We interpret this behavior through LayerRecall's two routing decisions. At the
beginning of a reappearance, the current representation may still be
ambiguous. As identity, body shape, and scene relations become more
discriminative, current-conditioned retrieval updates the query and can better
align it with the relevant historical chunk. The retrieved K/V states are then
injected only into memory-sensitive layers, while the remaining layers retain
the backbone's local attention pathway. This separation allows historical
appearance cues to revise a residual attribute mismatch without reconstructing
the motion, pose, or scene as a whole. LayerRecall does not explicitly detect
errors; rather, the correction emerges from repeatedly conditioning memory
access on the evolving current state and restricting where recalled evidence
enters the network.

\subsection{Complete Cross-Backbone Portability Results}
\label{sec:supp_cross_backbone_complete}

\begin{figure*}[!t]
\centering
\includegraphics[width=1\textwidth]{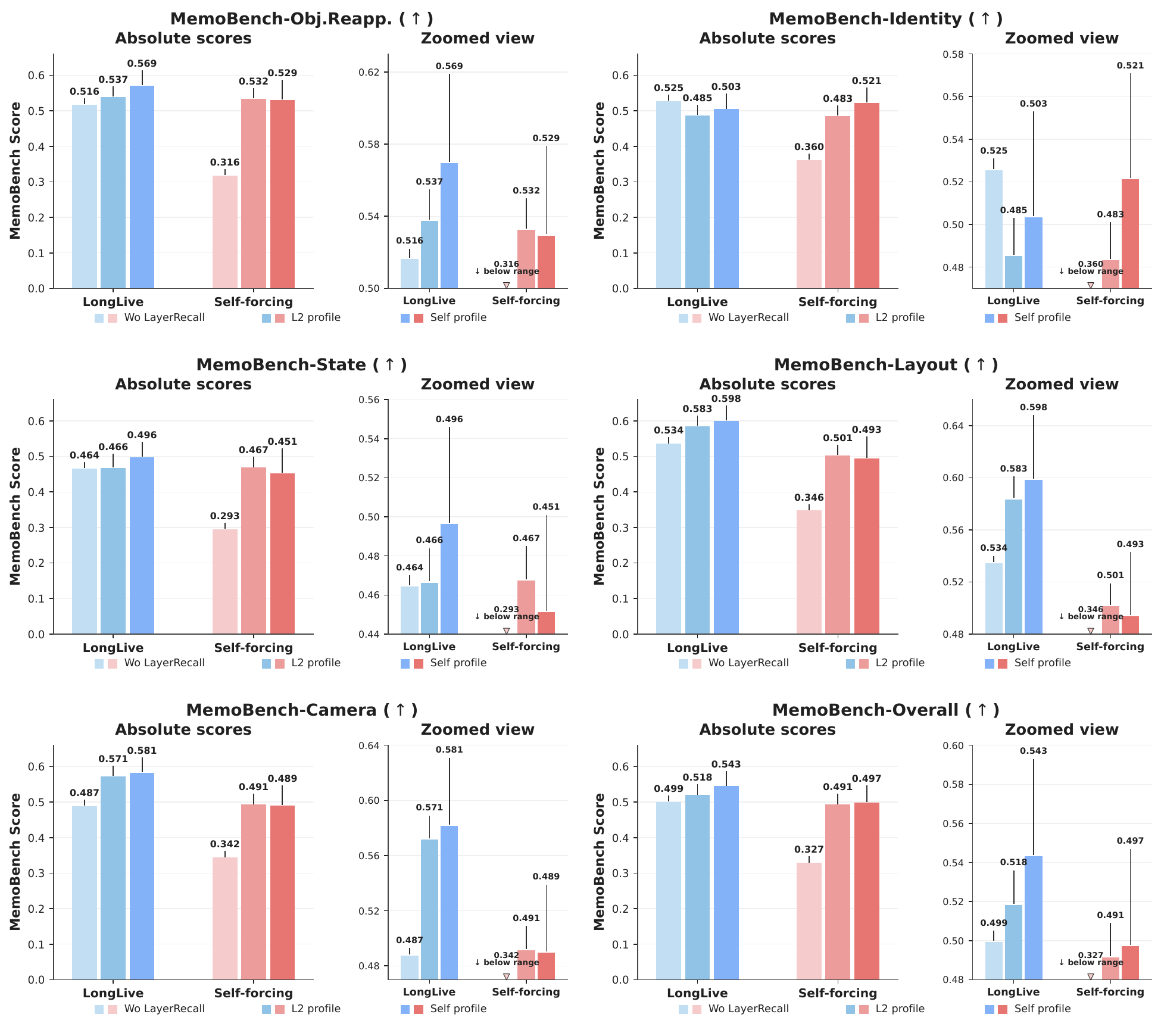}
\caption{Complete MemoBench results for zero-training LayerRecall transfer to
LongLive and Self-Forcing over 100 evaluation prompts. Each group compares the
backbone without LayerRecall, direct transfer using the LongLive-2.0 policy
(L2 profile), and the same frozen router using the target backbone's own
profiled insertion policy (Self profile). ``Self profile'' denotes
self-profiled layers for each target backbone, not Self-Forcing exclusively.
Absolute scores and zoomed views are shown for object reappearance, identity,
state, layout, camera, and overall performance. Thin black lines connect value
labels to their bars and do not denote uncertainty intervals.}
\label{fig:supp_cross_backbone}
\end{figure*}

Figure~\ref{fig:supp_cross_backbone} reports the complete MemoBench dimensions
for the cross-backbone portability experiment summarized in the main paper.
For both LongLive and Self-Forcing, it compares the frozen backbone, direct
LayerRecall transfer with the LongLive-2.0 layer policy, and transfer with only
the insertion layers replaced by the target backbone's profiled policy. These
results provide the fine-grained scores underlying the main-paper portability
analysis; the claim is restricted to the two evaluated target backbones.

\section{Additional Method Details}
\label{sec:supp_method_details}

This section provides architecture, retrieval, layer-policy, and CHPM training
details omitted from the main paper for space.

\subsection{LayerRecall Architecture}
\label{sec:supp_architecture}

LayerRecall augments a frozen chunk-autoregressive DiT with a small
state-conditioned memory router. Each generated sequence is partitioned into
prompt blocks and latent chunks. In the submitted LongLive-2.0 setting
\cite{chen2026longlive2,wanai2025wan22ti2v5b}, one training/evaluation rollout
uses 384 latent frames, organized as 48 chunks of 8 latent frames.

\paragraph{Router architecture and initialization.}
The router operates in one attention-head dimension, $d_q=128$. Its shared
global query is initialized from $\mathcal N(0,0.02^2)$. The
current-conditioned branch consists of a LayerNorm over the 3,072-dimensional
DiT state, a $3{,}072\!\rightarrow\!512\!\rightarrow\!128$ MLP with a SiLU
activation, and a $3{,}072\!\rightarrow\!1$ linear gate followed by a sigmoid.
The final MLP projection is zero-initialized, so the router initially reduces
to the global query. The residual scale $\alpha$ is learnable and initialized
to $0.1$, while each learnable layer scale $\gamma_l$ is initialized to one.
Cosine scores use a fixed temperature $\tau=1.0$; the temperature is not a
trained parameter.

\paragraph{Historical summaries.}
For each historical chunk and each DiT layer $l$, LayerRecall pools the
normalized pre-RoPE keys into a compact summary $m_i^l$. This summary is used
only for retrieval scoring. The physical cache separately stores the full
RoPE-applied key and value tensors for the same chunk. Thus $m_i^l$ determines
which resident record is retrieved, whereas self-attention consumes its full
cached K/V payload rather than the summary.

\paragraph{Current state.}
At the current chunk $t$, LayerRecall mean-pools the detached pre-attention
hidden state at layer $l$ into $c_t^l$ and sets
$z_t^l=\operatorname{LN}(c_t^l)$. The retrieval query is
\begin{equation}
q_t^l =
q_{\mathrm{global}} +
\alpha \gamma_l
\sigma(W_g z_t^l + b_g)\operatorname{MLP}(z_t^l).
\end{equation}
The global query, LayerNorm, MLP, gate, and $\alpha$ are shared across layers;
only $\gamma_l$ is layer-specific. The global query provides a shared retrieval
prior, and the gated residual rotates this prior according to the current
content at each layer. Current-state pooling is detached from the frozen
backbone activations, while gradients still update the router through the query
residual.

\paragraph{Parameter count.}
The LayerRecall overlay contains 11 trainable tensors and 1,648,416 parameters:
the global query, LayerNorm, two-layer MLP, gate, 30 layer scales, and one
global residual scale. This is approximately $0.033\%$ of the 5B backbone.
The video DiT backbone, text encoder, VAE, and teacher/student generators remain
frozen during CHPM training.

\subsection{Hard and Soft Retrieval}
\label{sec:supp_hard_retrieval}

Both retrieval forms start from the cosine compatibility between the current
query and each physically resident historical summary,
\begin{equation}
s_i^l = \cos(q_t^l, m_i^l).
\end{equation}
During CHPM training, hard and soft retrieval are coupled through a
straight-through estimator: the selected hard K/V payload determines the
numerical forward pass, while a temperature-scaled soft mixture provides the
differentiable backward path to the compatibility scores. The submitted
training and main-evaluation configurations retain the eight highest-scoring
candidates before retrieval and use $\tau=1.0$.

\paragraph{Hard retrieval.}
For each memory slot, LayerRecall selects the highest-scoring candidate that
has not already been used, $i^*=\arg\max_i s_i^l$, and inserts that record's
complete cached K/V payload into self-attention. When the visible budget
contains multiple memory slots, previously selected records are masked before
the next selection, yielding retrieval without replacement. Our
evaluation uses \texttt{st\_topk\_slots}; under no-gradient inference, this
follows the same hard numerical path without the soft gradient surrogate.

\paragraph{Soft retrieval.}
Over the retained candidate set $\mathcal C_l$, LayerRecall computes
\begin{equation}
w_i^l =
\frac{\exp(s_i^l/\tau)}{\sum_{j\in\mathcal C_l}\exp(s_j^l/\tau)},
\end{equation}
and forms $\widetilde K^l=\sum_i w_i^lK_i^l$ and
$\widetilde V^l=\sum_i w_i^lV_i^l$. During CHPM training, these mixtures
supply the backward surrogate through
\begin{equation}
\begin{aligned}
K_{\mathrm{ST}}^l
&=\operatorname{sg}\!\left(K_{i^*}^l-\widetilde K^l\right)+\widetilde K^l,\\
V_{\mathrm{ST}}^l
&=\operatorname{sg}\!\left(V_{i^*}^l-\widetilde V^l\right)+\widetilde V^l.
\end{aligned}
\end{equation}
The resulting tensors are numerically equal to the hard payload in the forward
pass, but their gradients reach the compatibility scores through the soft
mixture. Detached backbone summaries prevent this gradient from updating the
base DiT. The implementation also supports a pure-soft mode in which the
weighted payload itself enters attention, but this mode is not used for the
reported main evaluation.

\subsection{Layer Profiling and Layer Policy Selection}
\label{sec:supp_layer_policy}

The main paper states that the memory-sensitive layer set
$\mathcal L_{\mathrm{mem}}$ is chosen through attention profiling,
routing-usage analysis, and controlled policy comparisons. We use profiling to
identify layer-wise temporal heterogeneity, then fix a backbone-specific
insertion policy before evaluation.

\paragraph{Aggregation.}
For each layer, profiling first aggregates historical attention mass over query
tokens and heads, then averages it over denoising steps and query chunks to
obtain one profile per prompt. We report the prompt mean together with the
prompt-level min--max range. The compact visualization additionally groups
prompts by four memory stresses: subject return, color/attribute memory, object
permanence, and spatial memory. This aggregation
exposes stable layer-level trends without using any profiling prompt for CHPM
training or final evaluation.

\paragraph{Policy selection.}
Raw attention ranking is used to form candidate policies rather than directly
serving as the final insertion mask. We compare these candidates under a fixed
checkpoint, retrieval budget, and evaluation protocol, and then freeze the
selected backbone-specific allowlist. For LongLive-2.0, the submitted policy is
\begin{equation}
\mathcal L_{\mathrm{mem}} =
[4,9,10,12,13,15,16,17,18,26].
\end{equation}
Profiling therefore identifies where historical access is useful, while
controlled policy comparison determines the final insertion set; the allowlist
is not defined as a literal top-$k$ ranking of a single attention statistic.

\paragraph{Cross-backbone insertion policies.}
For the cross-backbone study, profiling is used only to choose where the
already-trained LongLive-2.0 LayerRecall router is inserted. We do not train
separate LongLive- or Self-Forcing-specific router parameters. The main
zero-training setting reuses the LongLive-2.0 insertion policy; its
backbone-profiled variant changes only the enabled-layer mask while keeping the
router weights fixed. The target-backbone insertion policies are
\[
\begin{aligned}
\mathrm{LongLive}\!:\;&[15,18,13,22,24,25,26,19,11,23],\\
\mathrm{Self\mbox{-}Forcing}\!:\;&[15,18,6,13,10,2,24,23,3,25].
\end{aligned}
\]
These layer policies are fixed before the final cross-backbone evaluation; all
LayerRecall rows reuse the same trained router weights without further
optimization.

\begin{table*}[t]
\centering
\small
\setlength{\tabcolsep}{5pt}
\begin{tabular}{l r p{0.34\textwidth} c c c}
\toprule
Split & Size & Purpose & CHPM train & Layer selection & Final eval \\
\midrule
Training set & 1,600 & Multi-shot prompts used to train CHPM. The prompt summary records 480 3-shot, 480 4-shot, 320 5-shot, and 320 6-shot cases across subject, color, position, and scene axes. & Yes & No & No \\
Profiling set & 100 & Held-out prompts used for prompt-stability analysis and layer-policy validation across subject, color, object, and spatial memory stresses. & No & Yes & No \\
Main evaluation set & 100 & Held-out 3-shot prompts used to generate the main evaluation videos. The summary records 25 cases for each of subject, color, position, and scene axes. & No & No & Yes \\
\bottomrule
\end{tabular}
\caption{Data split usage for training, layer-policy analysis, and evaluation.}
\label{tab:supp_splits}
\end{table*}

\subsection{CHPM Training Details}
\label{sec:supp_chpm_training}

CHPM trains only the LayerRecall router by matching a bounded-memory student to
a privileged long-context teacher in denoised-latent prediction space. Teacher
and student use the same LongLive-2.0/Wan2.2-TI2V-5B backbone checkpoint and
the same current noisy latent, text condition, and diffusion timestep at each
supervised anchor. Both teacher and student backbones are frozen. The complete
training configuration is summarized in Table~\ref{tab:supp_chpm_settings}.

\begin{table*}[t]
\centering
\small
\setlength{\tabcolsep}{4pt}
\begin{tabular}{l p{0.72\textwidth}}
\toprule
Item & Setting \\
\midrule
Backbone & LongLive-2.0 5B on Wan2.2-TI2V-5B. \\
Trainable parameters & LayerRecall router only; 1,648,416 parameters. \\
Training prompts & 1,600 multi-shot prompts. \\
Latent rollout & 384 latent frames, 48 chunks, 8 latent frames per chunk. \\
Prediction schedule & One sampled diffusion timestep and one denoising prediction-matching forward per supervised anchor. \\
Teacher/student history & Teacher retains the complete causal prefix at each anchor, with a maximum attention horizon of 384 latent frames; student has a 32-frame total attention-visible budget comprising an 8-frame sink, up to two retrieved history chunks (16 frames), and the current 8-frame chunk. \\
Anchors & Prediction anchor every 64 frames at chunk indices [7, 15, 23, 31, 39, 47]. \\
Loss & Mean-squared error between student and detached teacher denoised-latent predictions, with prediction loss weight 1.0 and regularization weight 0.0001. \\
Detached rollout & Student prefixes and anchor predictions are detached before becoming future context; teacher target predictions are detached. \\
Optimizer & AdamW, learning rate $1\times10^{-5}$, betas (0.0, 0.999), weight decay 0. \\
Batching & Per-DP batch size 1, gradient accumulation 1, effective global batch 8. \\
Precision & Mixed precision enabled for training; evaluated generator checkpoint uses BF16 inference. \\
Parallelism & 16 H100 GPUs, world size 16, sequence parallelism 2, data parallelism 8. \\
Schedule & One pass over 1,600 training cases, corresponding to 200 optimizer steps at data-parallel size 8. \\
Random seed & Logging, model initialization, and data-order seeds are all set to 0. \\
Training compute & 128 H100 GPU hours. \\
\bottomrule
\end{tabular}
\caption{CHPM training and reproducibility settings.}
\label{tab:supp_chpm_settings}
\end{table*}

\section{Prompt Construction, Baseline Adaptation, and Evaluation Protocol}
\label{sec:supp_prompt_adaptation}

\subsection{Prompt Construction}
\label{sec:supp_prompt_construction}

The CHPM training set contains 1,600 multi-shot prompts. The prompt summary
records 480 3-shot, 480 4-shot, 320 5-shot, and 320 6-shot cases. The four
paper axes are balanced at 400 prompts each: subject, color, position, and
scene. The corresponding primary stresses are subject return identity, color
consistency memory, spatial position memory, and scene layout consistency.

The main evaluation bank contains 100 held-out 3-shot prompts with 25 prompts
per paper axis. Each prompt is written as a sequence in which salient subjects,
attributes, objects, spatial relations, or scene landmarks appear, leave the
local context, and later reappear. Prompt metadata records target attributes,
bad-case criteria, camera views, shot durations, memory axes, and quality
control fields. A separate 100-prompt profiling bank is used for layer-policy
analysis and is disjoint from both CHPM training and the final 100-prompt
evaluation set.

\subsection{Prompt Filtering}
\label{sec:supp_prompt_filtering}

The checked prompt summaries report empty quality-issue distributions for the
1,600-prompt training bank and the 100-prompt main evaluation bank. Metadata
fields indicate filtering for target axis, overloaded consistency targets,
smooth transitions without explicit scene-transition wording, and limited
color clutter for non-color axes.

\subsection{Baseline Prompt Adaptation}
\label{sec:supp_baseline_adaptation}

Some baselines accept a native multi-shot or block-structured prompt, while
others are driven by a single text prompt. For single-prompt baselines, we
adapt the multi-shot prompt into one descriptive prompt that preserves the key
subject, attribute, object, reappearance, and camera-return constraints without
adding new semantic targets.

\begin{table*}[t]
\centering
\scriptsize
\setlength{\tabcolsep}{3pt}
\begin{tabular}{p{0.14\textwidth} p{0.13\textwidth} p{0.17\textwidth} p{0.33\textwidth} p{0.13\textwidth}}
\toprule
Method & Native multi-shot support & Input format & Adaptation rule  \\
\midrule
LongLive & Yes/partial & Block prompt list & Use the 48-block prompt script directly when supported. \\
LongLive-2.0 & Yes/partial & Block prompt list & Use the 48-block prompt script directly.  \\
LongLive-RAG & Partial & Retrieved prompt/context & Preserve the same shot script and use its retrieval interface.  \\
CausVid & No/partial & Single text prompt & Concatenate shots into a coherent long-video prompt with explicit reappearance constraints.  \\
Deep Forcing & No/partial & Single text prompt & Concatenate shots; preserve subject, layout, and return constraints.  \\
Dummy Forcing & No/partial & Single text prompt & Concatenate shots; preserve the memory target and final return.  \\
Self-Forcing & No/partial & Single text prompt & Use the single-prompt adaptation produced from the same shot script.  \\
MemFlow & No/partial & Single text prompt & Concatenate shots while preserving memory target and transition.  \\
Rolling Forcing & No/partial & Single text prompt & Concatenate shots into a long-video prompt. \\
Context Forcing & No/partial & Single text prompt & Concatenate shots into a long-video prompt.  \\
HiAR & No/partial & Single text prompt & Concatenate shots into a long-video prompt.  \\
Infinity-RoPE & No/partial & Single text prompt & Concatenate shots into a long-video prompt.  \\
SkyReels-V2 & No/partial & Single text prompt & Concatenate shots into a long-video prompt.  \\
LayerRecall & Yes & 48-block prompt list & Use the 48-block prompt script directly.  \\
\bottomrule
\end{tabular}
\caption{Baseline prompt adaptation policy. ``No/partial'' means the local
evaluation used a single-prompt adaptation or a method-specific wrapper rather
than assuming a native 48-block prompt interface.}
\label{tab:supp_baseline_adaptation}
\end{table*}

\paragraph{Adaptation example 1.}
Original 3-shot prompt: (1) a female archivist named Mara with a copper hairpin
and square glasses sits at a reading desk under a skylight; (2) the lens moves
away to a far bookshelf while Mara leaves the frame; (3) the camera returns to
the reading desk and Mara reappears with the same copper hairpin, glasses, and
grey cardigan. Adapted single prompt: generate a continuous long video in a
hushed library archive where Mara is introduced at the desk with the copper
hairpin and square glasses, the camera smoothly leaves her toward the far
bookshelf, and later returns to the desk where Mara reappears with the same
identity, accessories, clothing cues, and surrounding globe/skylight layout.

\paragraph{Adaptation example 2.}
Original 3-shot prompt: (1) a man in a charcoal suit stands in a train station
with a cognac-brown leather briefcase at his feet by a marble pillar; (2) the
camera moves toward the departure board and the briefcase leaves view; (3) the
camera returns and the same cognac-brown briefcase and station layout are
visible again. Adapted single prompt: generate a continuous long video inside a
grand train station where the cognac-brown leather briefcase is established
beside the man and marble pillar, temporarily leaves the local view as the
camera moves to the departure board, and later reappears with the same color,
material, owner/location relation, and station landmarks.

\subsection{Reproducibility and Evaluation Protocol}
\label{sec:supp_reproducibility}

\paragraph{Video checks and frame extraction.}
The trade-off evaluation script requires 100 generated videos before running
the full evaluation. It extracts inspection frames every 10 seconds up to 60
seconds, requires an expected minimum duration of 55 seconds, and uses ffprobe
and ffmpeg when available with an OpenCV fallback.

\paragraph{Metric execution.}
CLIP/DINO consistency, VBench, VBench-Long, Helios, MemoBench-style VLM
scoring, and MovieBench-style VLM scoring are run on all 100 evaluation
videos. VBench dimensions include temporal flickering, subject consistency,
background consistency, motion smoothness, aesthetic quality, and imaging
quality. VBench-Long dimensions include subject consistency, background
consistency, motion smoothness, dynamic degree, aesthetic quality, and imaging
quality.

\paragraph{Temporal stability diagnostic.}
For decoded frames, we convert each frame to grayscale, resize it to 192 pixels
in width, and define frame-change energy as
$e_t=\operatorname{mean}(|G_t-G_{t-1}|)$. Abrupt-change energy is the positive
residual $a_t=\max(e_t-\operatorname{MA}_{w}(e)_t,0)$, where
$w=\max(3,\operatorname{round}(2\,\mathrm{s}\cdot\mathrm{fps}))$ and the moving
average uses edge-value padding. For each matched video pair, we normalize both
abrupt-energy sequences by their shared 95th percentile before pointwise
averaging. We compute the temporal spectrum as
$P(f)=|\operatorname{rFFT}((e-\bar e)\odot\operatorname{Hann})|^2$ and report
$\sum_{2\leq f\leq12}P(f)/\sum_{0.2\leq f\leq12}P(f)$ as the high-frequency
ratio. Each displayed spectrum is normalized by its maximum power over
$0.05$--$12$ Hz, while the reported ratio is averaged over video-level ratios.

\paragraph{VLM protocol.}
All VLM-based evaluation uses \texttt{gemini-3-flash-preview}. For each video,
the evaluator samples 12 evenly spaced frames, resizes the long side to 512
pixels, and requests JSON scores under the MemoBench and MovieBench schemas.
Video-level scores are then averaged over all 100 evaluation prompts.

\paragraph{Aggregation.}
Reported table values are arithmetic means over the corresponding videos and
dimensions. The main VBench-Long Avg. is the mean of subject consistency,
background consistency, and motion smoothness. MemoBench and MovieBench
overall values are the evaluator-provided overall fields averaged over videos,
not the mean of only the dimensions selected for the main table.

\paragraph{Runtime.}
End-to-end time is measured around the generation subprocess. Decoded-frame
FPS is computed as decoded RGB frames divided by that end-to-end wall-clock
time. This is separate from playback FPS, which is only the MP4 playback rate.
The 48-block latent-FPS audit additionally reports latent frames per second as
384 latent frames per video divided by the same wall-clock interval; this
latent-FPS value is retained as an audit statistic and should not be mixed with
decoded-frame FPS.

\end{document}